\documentclass[10pt,twocolumn,letterpaper]{article}

\usepackage[pagenumbers]{cvpr}
\usepackage{xcolor}
\usepackage{arydshln}
\usepackage{multirow}
\usepackage{makecell}
\usepackage{graphicx}   
\usepackage{booktabs}   
\usepackage[table]{xcolor} 

\definecolor{cvprblue}{rgb}{0.21,0.49,0.74}
\usepackage[pagebackref,breaklinks,colorlinks,allcolors=cvprblue]{hyperref}

\def\paperID{*****} 
\def\confName{CVPR}
\def\confYear{2026}

\title{PCSTracker: Long-Term Scene Flow Estimation for Point Cloud Sequences}

\author{Min Lin$^{1}$, ~~Gangwei Xu$^{1}$, ~~Xianqi Wang$^{1}$, ~~Yuyi Peng$^{1}$,~~Xin Yang$^{2,1}$\footnotemark[2]\\
[2mm]
$^1$Huazhong University of Science and Technology \quad $^2$Optics Valley Laboratory\\}

\begin{document}
\maketitle
\begin{abstract}

Point cloud scene flow estimation is fundamental to long-term and fine-grained 3D motion analysis.
However, existing methods are typically limited to pairwise settings and struggle to maintain temporal consistency over long sequences as geometry evolves, occlusions emerge, and errors accumulate. In this work, we propose PCSTracker, the first end-to-end framework specifically designed for consistent scene flow estimation in point cloud sequences. Specifically, we introduce an iterative geometry–motion joint optimization module (IGMO) that explicitly models the temporal evolution of point features to alleviate correspondence inconsistencies caused by dynamic geometric changes. In addition, a spatio-temporal point trajectory update module (STTU) is proposed to leverage broad temporal context to infer plausible positions for occluded points, ensuring coherent motion estimation. To further handle long sequences, we employ an overlapping sliding-window inference strategy that alternates cross-window propagation and in-window refinement, effectively suppressing error accumulation and maintaining stable long-term motion consistency. Extensive experiments on the synthetic PointOdyssey3D and real-world ADT3D datasets show that PCSTracker achieves the best accuracy in long-term scene flow estimation and maintains real-time performance at 32.5 FPS, while demonstrating superior 3D motion understanding compared to RGB-D–based approaches. Code is available at https://github.com/MinLin2022/PCSTracker

\end{abstract}    
\section{Introduction}
\label{sec:intro}

Understanding long-term fine-grained 3D motion from point cloud sequences is fundamental to perceiving temporal dynamics in complex scenes.
Such temporally consistent motion estimation enables the recovery of detailed kinematic attributes, providing rich cues to interpret object–environment interactions and supporting advanced applications in autonomous driving~\cite{teng2023motion}, robotic navigation~\cite{noh20253d,duisterhof2023deformgs}, and virtual or augmented reality~\cite{dong2025dynamic,wang2023pointshopar}.

However, current approaches~\cite{giancola2019leveraging,qi2020p2b,chen2023tracking,liu2019flownet3d,wu2020pointpwc,wei2021pv}, following the two paradigms of object tracking and scene flow, remain inadequate for this goal. Object tracking~\cite{giancola2019leveraging,qi2020p2b,chen2023tracking} focuses primarily on motion at the object-level and fails to recover fine-grained motion. Meanwhile, despite significant advances in scene flow research for capturing fine-grained motion, existing methods~\cite{liu2019flownet3d,wu2020pointpwc,wei2021pv} remain largely limited to adjacent frames, making it difficult to maintain temporal consistency over long temporal spans.

Directly scaling short-term methods to long sequences ($e.g$, tens to hundreds of frames) by naively chaining predictions leads to catastrophic errors. Viewpoint changes and object deformations introduce substantial temporal dynamics in point features, undermining the consistency of point correspondences and resulting in inaccurate motion estimation. Frequent occlusions and out-of-bounds motions disrupt point correspondences, posing a fundamental challenge to maintaining temporal motion continuity. In addition, minor errors inevitably accumulate over time and eventually lead to severe drift.

To overcome these limitations, we present a comprehensive investigation into the largely unexplored area of long-term scene flow estimation for point cloud sequences, which can be regarded as both a temporal extension of two-frame scene flow and a point-level refinement of object tracking. We introduce PCSTracker, the first model that robustly and efficiently predicts long-term scene flow directly from raw point cloud sequences. Given a sequence of length $T$ point cloud frames and the initial position of arbitrary query points, our method outputs their complete 3D trajectories as a $T \times 3$ matrix.

PCSTracker addresses these core challenges through three key designs. First, an iterative geometry–motion joint optimization module (IGMO) refines query points’ geometry features and motions across frames based on local geometric similarity, explicitly modeling the temporal evolution of local geometry to preserve stable correspondences under dynamic changes.
Second, a spatial-temporal point trajectory update module (STTU) is employed to capture both inter-frame temporal dependencies and intra-frame spatial dependencies within a temporal window. It leverages sparse visible timesteps and temporal context to infer plausible positions in intermediate frames under occlusions or out-of-bound motion, thereby enhancing robustness against correspondence loss and ensuring motion continuity across frames. 
Finally, an overlapping sliding-window strategy is adopted, where adjacent windows partially overlap, and the model alternates cross-window prediction propagation with in-window refinement. This design enforces temporal consistency across windows and effectively suppresses error accumulation and motion drift in long sequences.

For training and evaluation, we construct two datasets: a large-scale synthetic dataset, PointOdyssey3D, derived from PointOdyssey~\cite{zheng2023pointodyssey}, and a real-world dataset, ADT3D, built upon the Aria Digital Twin (ADT)~\cite{pan2023aria} for generalization testing. Extensive experiments show that our method achieves superior temporal consistency and robustness over point-cloud-based pairwise methods with naive chaining, reducing $\text{EPE}_{3D}$ by 57.9\% on PointOdyssey3D and 60.6\% on ADT3D, while achieving high computational efficiency with an inference speed of 32.5 FPS. Visualization results in Fig.~\ref{fig:visualization of track} further show that our method provides more accurate 3D motion understanding than RGB-D–based approaches. In summary, our main contributions are as follows:

\begin{enumerate}
    \item This work presents the first exploration of long-term scene flow estimation in point clouds, enabling fine-grained point-level motion reasoning over extended temporal spans in raw point cloud sequences.

    \item We introduce three key designs that jointly enforce temporal consistency by mitigating the effects of dynamic geometric changes, occlusions, and error accumulation.

    \item We construct two benchmark datasets, PointOdyssey3D (synthetic) and ADT3D (real-world), for training and evaluation, serving as resources for long-term point cloud scene flow research.
    
    \item  Our method achieves state-of-the-art performance on both datasets, significantly outperforming previous approaches while maintaining high inference speed.
    
\end{enumerate}


\section{Related Work}
\label{sec:Related}

\subsection{Point Cloud-based Scene Flow Estimation}

Scene flow estimation traces its origins to the seminal work of Vedula et al.\cite{vedula1999three}, with early research focusing on RGB-based methods \cite{mehl2023m,schuster2020sceneflowfields++,liu2024joint,jaimez2015motion,ma2019deep}. With the development of point cloud processing techniques~\cite{qi2017pointnet,qi2017pointnet++,wu2019pointconv,wu2023pointconvformer}, the field has gradually shifted toward end-to-end frameworks that estimate scene flow directly from point cloud data~\cite{liu2019flownet3d,gu2019hplflownet,puy2020flot,ouyang2021occlusion,wang2021festa,cheng2023multi,liu2024difflow3d}.
FlowNet3D~\cite{liu2019flownet3d} pioneered end-to-end point cloud scene flow estimation with an encoder–decoder architecture, introducing flow embedding and set upconv layers to correlate geometric features across frames. PointPWC-Net~\cite{wu2020pointpwc} adopted a coarse-to-fine design and a learnable cost volume layer that efficiently aggregates local motion cues without heavy 4D tensors. Later works~\cite{cheng2022bi,wang2022matters} refined this direction through bidirectional aggregation and backward reliability validation. PV-RAFT~\cite{wei2021pv} advanced the field by introducing an iterative refinement architecture that combines point-wise and voxel-wise correlation volumes to construct a pyramid correlation field, together with a GRU-based iterative update module and a learnable flow refinement module. Later methods~\cite{fu2023pt,cheng2023multi,liu2024difflow3d,lin2025flowmamba} were developed upon this framework, emphasizing more efficient correlation modeling and more robust inference in complex and dynamic scenarios. However, these pairwise methods are inherently limited for long-term tracking. In contrast, our approach can predict scene flow across tens or even hundreds of frames.

 \subsection{ Object Tracking in Point Clouds}

The importance of leveraging broad temporal context has been previously explored in point cloud object tracking~\cite{giancola2019leveraging,qi2020p2b,wang2025trackany3d,zhang2019frame,weng20203d,weng2021captra,sun2024l4d,chen2023tracking}. This task can be broadly categorized into two main types: trajectory-based tracking, which focuses on following the positions of individual or multiple objects over time, and pose tracking, which additionally estimates the objects' orientations or full 6-DoF poses. Trajectory-based tracking methods can be further distinguished by the number of targets: single-object tracking (SOT)~\cite{qi2020p2b, fang20203d, giancola2019leveraging, zhang2024robust, zheng2021box, lin2024seqtrack3d, nie2025p2p, ma2023synchronize, wang2023correlation}, which follows a single target over time often from a given initial position, and multi-object tracking (MOT)~\cite{weng20203d, zhang2019frame, zhu2018online, milan2013continuous, dehghan2015gmmcp, tang2016multi, liang2020pnpnet}, which simultaneously tracks multiple objects while addressing challenges such as identity switches and data association. Pose tracking methods~\cite{weng2021captra,chen2023tracking, liu2022catre, sun2024l4d, zhao2023learning, zhang2023generative, zheng2023hs} primarily focus on rigid or articulated objects with known shapes. However, these approaches are not tailored for deformable or weakly textured objects and provide limited information about surface deformation. In contrast, our approach extends object tracking to the point level, enabling fine-grained motion estimation

\begin{figure*}[t]
    \centering
    \includegraphics[width=1.0\linewidth]{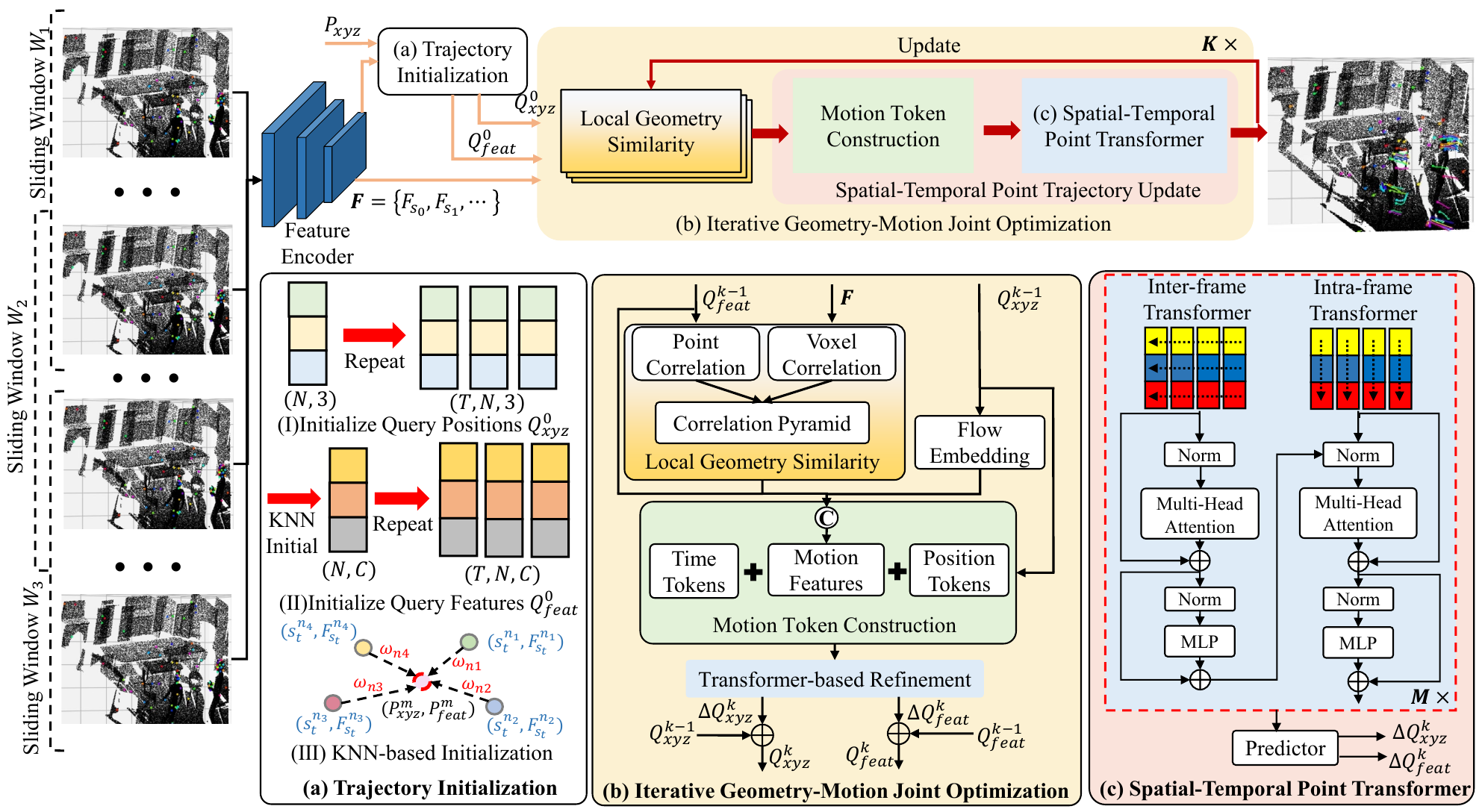}
    \caption{\textbf{Overview of PCSTracker pipeline.} Given a point cloud sequence, we first divide it into partially overlapping sliding windows. Within each window, query points are initialized via a KNN-based interpolation to form initial trajectories. Local geometric similarities are then computed through point–voxel correlations and processed by a spatio-temporal point Transformer to capture long-range spatial–temporal dependencies. Finally, trajectories are iteratively refined through $K$ optimization steps.}
    \label{fig:overview network}
    \vspace{-10pt}
\end{figure*}

\subsection{Long-Term Scene Flow Estimation}

Current approaches to long-term scene flow estimation predominantly rely on 2D representations as input. The work in~\cite{wang2025scenetracker} first introduced the long-term scene flow estimation (LSFE) task, which leverages RGB-D data to construct joint appearance–depth features through dynamic indexing, and employs a Transformer-based iterative framework to separately optimize 2D trajectories and depth. DeformGS~\cite{duisterhof2023deformgs} relies solely on RGB inputs and adopts a 4D Gaussian representation combined with neural voxel encoding, deformation learning, and physical regularization to achieve long-term scene flow estimation. However, due to the intrinsic differences between point clouds and image modalities, these methods cannot be directly applied when only point clouds are available. Designing a long-term scene flow estimation framework tailored to irregular point cloud data is the central focus of our research.


\section{Methods}
\label{sec:method}

Our goal is to predict long-term scene flow throughout point cloud sequences. Formally, point cloud sequences of length $T$ are represented as
$\mathbf{S} = \{S_t \in \mathbb{R}^{N_t \times 3}\}_{t=1}^T$, 
where each frame contains $N_t$ points and the $j$-th point at time $t$ is denoted as $S_t^j$. 
The proposed PCSTracker predicts complete temporal trajectories for $N$ query points with given initial coordinates, represented as ${P}_{xyz} \in \mathbb{R}^{N \times 3}$. By default, the query coordinates originate from the first frame of the sequence; however, our method also supports flexible initialization from any intermediate frame.

The overall pipeline of the proposed method is illustrated in Fig.~\ref{fig:overview network}. We first describe the feature extraction \& trajectory initialization process in Sec. \ref{FETI}. Subsequently, we detail the iterative geometry-motion joint optimization module (IGMO) in Sec. \ref{IGMO}. Next, we introduce the spatial-temporal point trajectory update module (STTU) in Sec. \ref{FET}. Finally, the window-based inference and training are discussed in Sec .~\ref {WITS}.

\subsection{Feature Extraction \& Trajectory Initialization}
\label{FETI}

\noindent\textbf{Feature Extraction} For each input frame $S_t$, we utilize PointConv as the backbone network to extract $C$-dimensional feature representations, denoted as ${F}_{s_t} \in \mathbb{R}^{{N_t} \times C}$. To enhance computational efficiency, the input point cloud is hierarchically downsampled via farthest point sampling (FPS). Specifically, the feature encoder contains two FPS-based downsampling layers that progressively capture local geometric relationships at multiple scales, resulting in an overall spatial reduction factor of $s=4$. The encoder finally outputs the feature map $\mathbf{F} = \{F{s_t} \in \mathbb{R}^{N_t/4 \times C}\}_{t=1}^T$.

\noindent\textbf{Trajectory Initialization} 
We initialize the position and features of each trajectory based on the position and features of query points, denoted as ${Q_{xyz}^0 \in \mathbb{R}^{T\times{N} \times C}}$ and ${Q_{feat}^0 \in \mathbb{R}^{T\times{N} \times C}}$, respectively. Specifically, as shown in Fig. \ref{fig:overview network} (a), the feature of each query point is obtained through $K$-nearest-neighbor (KNN) interpolation from the feature maps $\mathbf{F}$ and replicated across all timesteps to initialize the trajectory state.

For each query ${P_{xyz}^m}$ from t-th frame $S_t$, we find its $N_k$ nearest neighbors, with Euclidean distances $d_j = \|{P_{xyz}^m} - {S_t}^j\|_2$. The interpolated feature is then computed as:

\begin{equation}
P_{feat}^m = \sum_{j=1}^{N_k} w_j \, F{s_t}^j, 
\quad 
w_j = \frac{1}{d_j } \Bigg/ \sum_{j=1}^{N_k} \frac{1}{d_j},
\end{equation}

It is worth noting that we allow query points to be placed at arbitrary positions, not limited to the scanned points, but also including physically meaningful locations that are not directly captured by the 3D sensor.

\subsection{Iterative Geometry-Motion Joint Optimization}
\label{IGMO}

Geometry evolution breaks the consistency of correspondences, making it difficult to maintain reliable motion estimation in dynamic scenes. To address this issue, we explicitly model the temporal variations of query points and progressively refine their geometry and motion.
As illustrated in Fig.~\ref{fig:overview network} (b), the module takes as input the trajectory positions $Q_{xyz}^{k-1}$ and features $Q_{feat}^{k-1}$ from the previous iteration, together with the point cloud feature maps $\mathbf{F}$. It then outputs the updated positions $\Delta Q_{xyz}^{k}$ and features $\Delta Q_{feat}^{k}$ of the trajectory for the next iteration.

Specifically, for the $k$-th iteration, we first measure the local geometric similarity ${C_{g}^{k} \in \mathbb{R}^{T\times{N} \times {N_t/4}}}$ between the updated features ${Q_{feat}^{k-1} \in \mathbb{R}^{T\times{N} \times C}}$ of the trajectories from the previous iteration and the pre-computed feature maps ${\mathbf{F}\in \mathbb{R}^{T\times{N_t/4} \times C}}$ of the corresponding timestep. For memory efficiency and faster computation, a truncated correlation volume is further constructed by selecting the top-$M$ highest correlations in point cloud sequences $\mathbf{S}$, denoted as ${C^{k} \in \mathbb{R}^{T \times N \times M}}$.  Following the PV-RAFT \cite{wei2021pv}, we adopt a dual-branch correlation module to construct the correlation pyramid for various time steps. 

\noindent\textbf{Point Correlation Branch}  
For each position of trajectory in $Q_{xyz}^k \in \mathbb{R}^3$, we select the top-${M_k}$ neighbors $\mathcal{N}_{M_k} = \mathcal{N}(S)_{M_k}$ in truncated point cloud sequences. The corresponding correlation values are denoted as ${C^{k}(\mathcal{N}_{M_k})}$. 
The fine-grained correlation features are formed by concatenating similarity scores with relative positional offsets, followed by feature aggregation.
\begin{equation}
C_{point}^k = \max\big( \text{MLP}\big( \text{concat}(C^k(\mathcal{N}_{M_k}), \mathcal{N}_{M_k} - Q_{xyz}^k) \big) \big)
\end{equation}

\noindent\textbf{Voxel Correlation Branch}  
The local 3D space around each trajectory position in $Q_{xyz}^k \in \mathbb{R}^3$ is discretized into $a \times a \times a$ cubes of varying sizes $r$. By setting varying $r$, correlation features at different spatial scales are constructed. For each cube, the points $\mathcal{N}_{r, a}^{a_i}$ in truncated point cloud sequences that fall inside the cube are identified, and their correlation values are averaged to generate the sub-cube features. 

\begin{equation}
    C^{k,r,a_i}_{subcub} = \frac{1}{|\mathcal{N}_{r,a}^{a_i}|} \sum C^k(\mathcal{N}_{r,a}^{a_i})
\end{equation}

Accordingly, the long-range correlation for a given $r$ is obtained by concatenating the correlation from the different sub-cubes and aggregating them through an MLP.

\begin{equation}
    C_{voxel}^{k,r} = \text{MLP}\big( \underset{a_i}{\text{concat}}(C^{k,r,a_i}_{subcub}) \big)
\end{equation}

Subsequently, the correlation features extracted from the two branches are fused into $C_{fuse}^k$ and fed into the spatial-temporal point trajectory update module. This module updates both motion and geometric features through a two-step process: motion token construction and spatial-temporal point Transformer. The updated outputs are then added to those from the previous iteration and obtain the refined motion and geometric features. The detailed designs of these two steps are presented in the following sections.

\subsection{Spatial-Temporal Point Trajectory Update}
\label{FET}

Frequent occlusions cause severe ambiguity in point correspondences, resulting in discontinuous motion. To address this, we estimate the complete motions of all query points jointly over the temporal span $T$, rather than independently at each frame. 
Leveraging a broader temporal context, the module infers plausible intermediate positions from sparse visible timesteps, maintaining temporal continuity and physical consistency in the trajectories.

In detail, we exploit the global modeling capability of the Transformer to capture long-range dependencies of query point motions across both temporal and spatial dimensions, improving estimation accuracy and robustness to occlusions. The module mainly consists of two parts: 1) Motion Token Construction and 2) Spatial-Temporal Point Transformer.

\subsubsection{Motion Token Construction}
As shown in Fig. \ref{fig:overview network} (b). The construction of motion tokens integrates correlation volumes $C_{fuse}^k$, trajectory features $Q_{feat}^{k-1}$, flow information, trajectory positions $Q_{xyz}^{k-1}$, and timestamps ${t}$. First, the flow information is embedded using a sinusoidal encoder $\eta_f$ and concatenated with the trajectory features $Q_{feat}^{k-1}$ and correlation volumes $C_{fuse}^k$ to form the motion features. Next, the positions and timestamps are encoded using $\eta_p$ and $\eta_t$, respectively, and added to the motion features, producing the final motion tokens. Formally, this process can be expressed as follows:

\begin{equation}
\mathbf{F}_{{motion}} = \text{Concat}(C_{fuse}^k, {Q_{feat}^{k-1}}, {\eta_f}({Q_{xyz}^{k-1}} - {Q_{xyz}^{0}}))
\end{equation}
\begin{equation}
\mathbf{F}_{{token}} = \mathbf{F}_{{motion}} + {\eta_p}({Q_{xyz}^{k-1}})+{\eta_t}({t})
\end{equation}

\subsubsection{Spatial-Temporal Point Transformer}
As shown in Fig. \ref{fig:overview network} (c). The motion tokens $\mathbf{F}_{{token}}$ are fed into a stack of $2 \times M$ Transformer blocks, where each Transformer consists of multi-head self-attention, normalization, and MLP layers. The spatial-temporal point Transformers alternately perform attention operations along the inter-frame and intra-frame dimensions, capturing global temporal and spatial dependencies to optimize these tokens.  

The optimized motion tokens $\mathbf{F}_{{token}}^o$ are then processed by a predictor $\Psi$ to estimate the residual motions and feature updates of the query points across all frames $(\Delta {Q_{xyz}^{k}}, \Delta {Q_{feat}^{k}}) = \Psi(\mathbf{F}_{{token}}^o)$.
These predictions are used to refine the trajectory positions and features for the next iteration:

\begin{align}
Q_{xyz}^{k} &= Q_{xyz}^{k-1} + \Delta Q_{xyz}^{k}, \\
Q_{feat}^{k} &= Q_{feat}^{k-1} + \Delta Q_{feat}^{k}.
\end{align}

\begin{figure}[!t]
    \centering
    \includegraphics[width=1.0\linewidth]{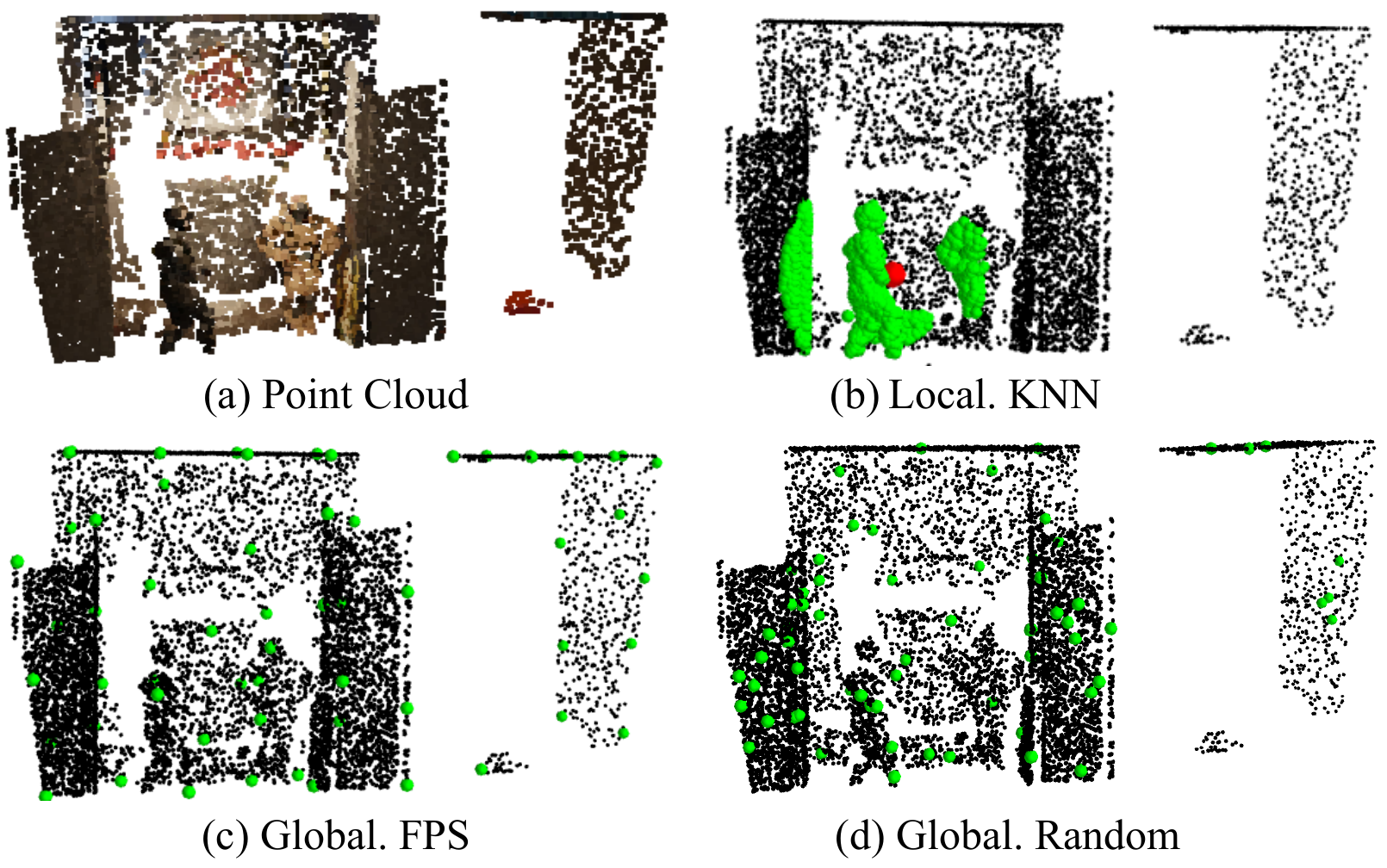}
    \caption{ \textbf{Auxiliary Points.} We visualize the auxiliary points selected by different sampling strategies, and Tab. \ref{tab:inference_mode} reports their impact on inference performance. \textcolor{green!50!gray}{Green points} denote the selected auxiliary points; (a) shows point clouds colorized with RGB information, and (b) highlights one of the query points in \textcolor{red}{red}.
    }
    \label{fig:support point}
    \vspace{-10pt}
\end{figure}

\subsection{Window-based Inference and Training}
\label{WITS}

For long point cloud sequences of total length $T'$ ($T' > T$), we adopt an overlapping sliding-window strategy. The sequence is divided into $W_{all} = \lceil 2T'/T - 1 \rceil$ sub-sequences of length $T$, with an overlap of $T/2$ between consecutive windows. Each window is initialized with the trajectory estimates from the previous one and optimized for $K$ iterations in the current window. The windows are processed sequentially, enforcing temporal continuity throughout the sequence and effectively reducing error propagation and motion drift over long time spans.

Additionally, PCSTracker supports different inference modes and allows the incorporation of either local or global auxiliary points, which have been shown~\cite{wang2024scenetracker,karaev23cotracker} to enhance trajectory estimation during inference. Due to the discrete and irregular nature of point clouds, auxiliary points cannot be directly constructed on a regular grid. Hence, as illustrated in Fig.~\ref{fig:support point}, we select local auxiliary points via K-Nearest Neighbors (KNN) and global ones via farthest point sampling (FPS) or random sampling.

We train our network in an unrolled manner to effectively manage semi-overlapping sliding windows. Given the 3D ground-truth positions $Q_{xyz}^{GT}$ within each window, the loss function is defined as follows, where we omit the window indices for simplicity.

\begin{equation}
Loss = \sum_{w=0}^{W_{all}} \sum_{t=1}^{T} \sum_{k=1}^{n} \gamma^{n-k} 
{
\| Q_{xyz}^{k,t,w} - Q_{xyz}^{GT} \|_2 
}
\end{equation}

where $Q_{xyz}^{GT}$ denotes the ground-truth positions of the query points, $\gamma$ is set to 0.8.

\section{Experiments}
\label{sec:experiments}

\subsection{Datasets for Training and Evaluation}

\noindent\textbf{PointOdyssey3D} Since there is no large-scale training dataset available for long-term scene flow estimation on point clouds, and obtaining scene flow annotations in real-world scenarios is extremely difficult and costly, we resort to a synthetic but challenging large-scale dataset to train and evaluate our model as well as to validate our design choices. Specifically, we construct a dataset named PointOdyssey3D based on the existing PointOdyssey dataset~\cite{zheng2023pointodyssey}, tailored for long-term scene flow estimation on point clouds while also reducing the data loading time. We generate point clouds by projecting pixels from RGB-D frames into the camera coordinate system using camera intrinsics. To mimic the sparsity and non-correspondence of real-world point clouds, we apply random sampling on the projected points. Each training sample consists of 24 consecutive point cloud frames, while each testing sample contains 40 frames, with 8,192 randomly sampled points per frame. To ensure sufficient supervision, we enforce that each sample contains at least 1,024 trajectories. In total, the dataset contains 32,307 training samples and 142 test samples.

\begin{table*}[t]
	\centering
	
	\label{tab: metric_3d}
	\resizebox{0.90\textwidth}{!}{%
		\begin{tabular}{cccccccccc}
			\toprule
			Method &Input & $\text{EPE}_{3D}\textcolor{red}{\downarrow}$ & $\delta_{3D}^{0.10}\textcolor{green!60!gray}{\uparrow}$ & $\delta_{3D}^{0.20} \textcolor{green!60!gray}{\uparrow}$ & $\delta_{3D}^{0.40} \textcolor{green!60!gray}{\uparrow}$ & $\delta_{3D}^{0.80}\textcolor{green!60!gray}{\uparrow}$ & $\delta_{3D}^{avg} \textcolor{green!60!gray}{\uparrow}$ & $\text{Survival}_{3D}^{0.50} \textcolor{green!60!gray}{\uparrow}$ & $\text{MAE}_{3D}\textcolor{red}{\downarrow}$  \\ \midrule
			SpatialTracker~\cite{SpatialTracker} & RGB-D & 0.924 & 21.12 & 33.03 & 48.06 & 66.81 & 42.25 & 49.54& 0.896 \\
            DELTA~\cite{ngo2024delta}  & RGB-D & 0.780 & 40.43 & 55.00 & 65.89 & 74.67 & 58.99 & 67.16& 0.762 \\
			SceneTracker~\cite{wang2025scenetracker} & RGB-D & 0.204 & 61.95 & 75.93 & 86.18 & 93.85 & 79.48 & 87.98 & 0.200 \\
            \midrule
            SF-baseline & Point & 0.330 & 28.92 & 51.77 & 74.88 & 91.03 & 61.65 & 77.78 & 0.319 \\
			PCSTracker(Ours) & Point & \textbf{0.133} & \textbf{70.19} & \textbf{84.48} & \textbf{93.37} & \textbf{97.44} & \textbf{86.37} & \textbf{93.65} & \textbf{0.129} \\

            \bottomrule
		\end{tabular}%
	}
    \caption{
        \textbf{Comparison results on the PointOdyssey3D.} We re-evaluate SpatialTracker~\cite{SpatialTracker}, SceneTracker~\cite{wang2025scenetracker}, and DELTA~\cite{ngo2024delta} on the corresponding RGB-D sequences of the PointOdyssey3D dataset using their publicly available code and pretrained checkpoints. In addition, we retrain the SF-baseline on PointOdyssey3D for a fair comparison. Our method achieves a reduction of 34.8\% and 59.7\% in $\text{EPE}_{3D}$ compared with SceneTracker and SF-baseline, respectively, demonstrating a substantial improvement in the estimation performance.		 
	}
    \label{tab:results of pointodyssey3d}
\end{table*}

\begin{figure*}[t]
    \centering
    \includegraphics[width=1.0\linewidth]{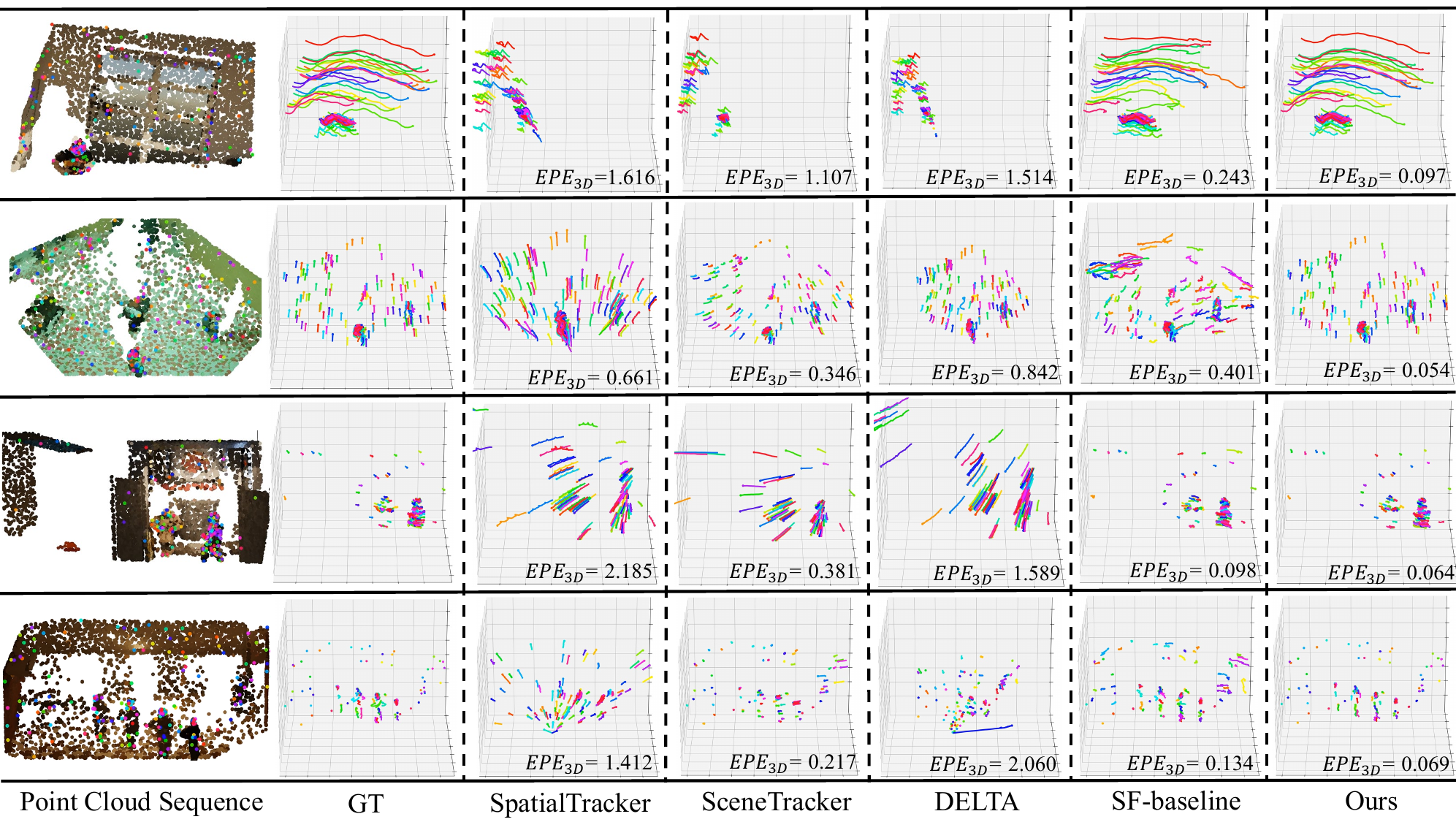}
    \caption{ \textbf{Qualitative Results on PointOdyssey3D.} The first column shows the input point cloud sequence (colored by RGB for visualization) and the corresponding query points. Columns 2–7 compare ground-truth trajectories with predictions from different methods. 
    RGB-D-based methods~\cite{SpatialTracker, wang2025scenetracker, ngo2024delta}, constrained by their 2D appearance-driven frameworks, fail to recover reliable 3D trajectories, reflecting limited understanding of scene geometry and 3D motion. 
    SF-baseline, which lacks a dedicated design for long sequences, suffers from significant errors. In contrast, our method produces more correct 3D motion understanding and accurate 3D trajectories.   
    }
    \label{fig:visualization of track}
    \vspace{-10pt}
\end{figure*} 

\noindent\textbf{ADT3D} We construct ADT3D, the first real-world evaluation dataset for this task, built upon the Aria Digital Twin (ADT) dataset~\cite{pan2023aria} released by Meta Reality Labs. The original ADT dataset comprises 200 sequences captured in two real indoor environments, containing 398 object instances (324 stationary and 74 dynamic) with synchronized depth maps, full camera intrinsics and extrinsics, and rich semantic annotations.
We select a subset of sequences, and using the provided depth maps and camera parameters, generate sparse point cloud sequences via random spatial sampling. By further integrating the 3D trajectory annotations from TAPVid-3D~\cite{koppula2024tapvid}, we construct a real-world benchmark of 498 sequences, each with 150 frames, for quantitative evaluation of long-term scene flow estimation.

\begin{table*}[t]
	\centering

	\resizebox{0.90\textwidth}{!}{%
		\begin{tabular}{cccccccccc}
			\toprule
			Method &Input & $\text{EPE}_{3D}\textcolor{red}{\downarrow}$ & $\delta_{3D}^{0.10}\textcolor{green!60!gray}{\uparrow}$ & $\delta_{3D}^{0.20} \textcolor{green!60!gray}{\uparrow}$ & $\delta_{3D}^{0.40} \textcolor{green!60!gray}{\uparrow}$ & $\delta_{3D}^{0.80}\textcolor{green!60!gray}{\uparrow}$ & $\delta_{3D}^{avg} \textcolor{green!60!gray}{\uparrow}$ & $\text{Survival}_{3D}^{0.50} \textcolor{green!60!gray}{\uparrow}$ & $\text{MAE}_{3D}\textcolor{red}{\downarrow}$  \\ \midrule
			SpatialTracker~\cite{SpatialTracker} & RGB-D & 0.916 & 23.71 & 38.04 & 58.16 & 80.29 & 50.05 & 60.31& 0.505 \\
            DELTA~\cite{ngo2024delta}  & RGB-D & 1.509 & 45.53 & 70.89 & 81.67 & 89.59 & 71.92 & 83.41 & 1.278 \\
			SceneTracker~\cite{wang2025scenetracker} & RGB-D & 0.601 & 43.15 & 62.44 & 79.46 & 90.91 & 68.99 & 80.40 & 0.271 \\
            \midrule
            SF-baseline & Point & 0.945 & 12.32 & 29.52 & 49.13 & 70.99 & 40.49 & 51.61 & 0.702 \\
			PCSTracker(Ours) & Point & \textbf{0.372} & \textbf{46.73} & \textbf{71.74} & \textbf{85.49} & \textbf{93.81} & \textbf{74.44} & \textbf{87.74} & \textbf{0.226} \\ 
            
            \bottomrule
		\end{tabular}%
	}
    \caption{
		\textbf{Generalization results on ADT3D dataset.}  We evaluate our method, SF-baseline, and RGB-D-based approaches on the ADT3D dataset and its corresponding RGB-D sequences. Our method reduces the $\text{EPE}_{3D}$ by 38.1\% and 60.6\% compared with SceneTracker~\cite{wang2025scenetracker} and SF-baseline, respectively, demonstrating its superior generalization ability and robustness in realistic dynamic scenes.
	}
    \label{tab: generalization}
\end{table*}

\begin{figure*}[t]
    \centering
    \includegraphics[width=1.0\linewidth]{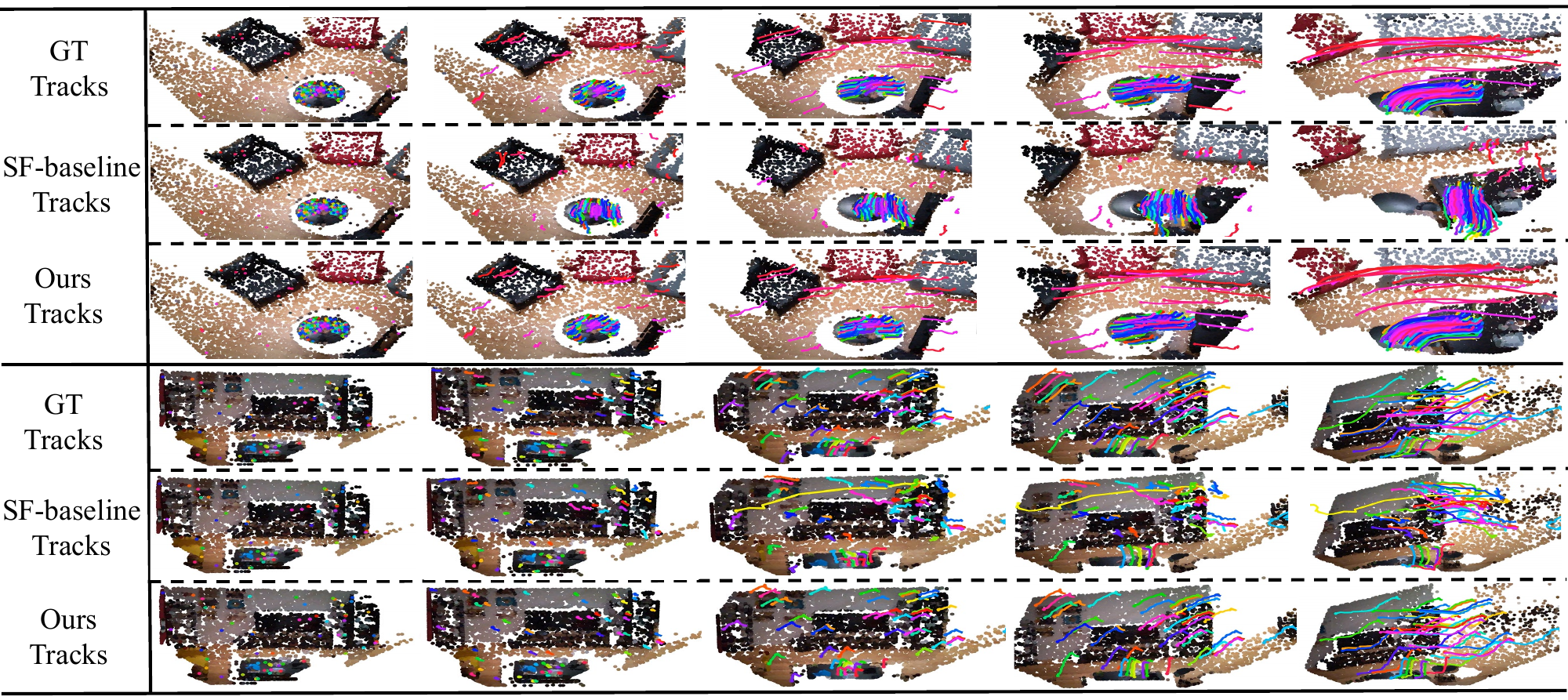}
    \caption{\textbf{Visualization of generalization results on ADT3D.} Each row shows the point clouds and the motion trajectories of the query points at different timestamps within a given scene (each point cloud is colored with its RGB information for better visualization). More results are provided in the appendix. }
    \label{fig:generalization on adt3d}
    \vspace{-10pt}
\end{figure*}

\subsection{Implementation Details and Metrics}

\noindent\textbf{Implementation Details} Our implementation is based on PyTorch, and we adopt AdamW as the optimizer with the OneCycle learning rate scheduling strategy. The model is trained on the PointOdyssey3D dataset with an initial learning rate of $2\times10^{-4}$ and a batch size of 4. Each training sample contains 24 point cloud frames and 256 query points, where each frame consists of 8192 points. We train the model for 200K steps by default. 
To construct the correlation volume pyramid, we set $r=\{0.25, 0.5, 1\}$. The number of $M$ is set to $3$. The number of iteration steps during both training and inference is set to 4.

\noindent\textbf{Metrics} We adopt the evaluation metrics used in recent studies~\cite{wang2024scenetracker}. Specifically, 
$\delta^x_{3\mathrm{D}}$ denotes the percentage of trajectory points whose 3D positional error is below a threshold distance $x \in \{0.10, 0.20, 0.40, 0.80\}$ meters from the ground truth. 
The average 3D position accuracy, $\delta^{avg}_{3D}$, is computed as the mean of $\delta^x_{3D}$ across all thresholds. In addition, we employ the median 3D trajectory error (MAE$_{3D}$) and the 3D end-point error (EPE$_{3D}$) to measure the discrepancy between the predicted and ground-truth trajectories. Finally, we use the $\text{Survival}_{3D}^{0.50}$ metric, where a trajectory is considered a failure if its error exceeds $0.50\,\mathrm{m}$.

\subsection{Method Comparisons}

We use PV-RAFT~\cite{wei2021pv} to construct a chain-based long-term baseline (SF-baseline) and retrain it on PointOdyssey3D for a fair comparison.
The query points from the first frame are propagated through the network to predict their positions in the next frame, and the predicted results are then used as queries for subsequent frames in sequence.
For SpatialTracker~\cite{SpatialTracker}, SceneTracker~\cite{wang2025scenetracker}, and DELTA~\cite{ngo2024delta}, we use the corresponding RGB frames and ground-truth depth videos from PointOdyssey3D as inputs.
As shown in Tab.~\ref{tab:results of pointodyssey3d}, our model achieves significant improvements in 3D accuracy. It reduces $\text{EPE}_{3D}$ by 34.9\% compared with SceneTracker, 85.6\% with SpatialTracker, 73.8\% with DELTA, and 59.7\% with SF-baseline. In Tab.~\ref{tab: generalization}, trained solely on synthetic data, it also generalizes well to real-world scenarios, lowering $\text{EPE}_{3D}$ by 38.1\% and 60.6\% compared with RGB-D-based methods and SF-baseline, respectively.
Qualitative results in Fig.~\ref{fig:visualization of track} show that our approach captures scene geometry and 3D motion more accurately than RGB-D-based methods and SF-baseline. It also demonstrates stronger robustness to occlusions and maintains more stable and temporally consistent trajectories than SF-baseline.
Further visualizations on ADT3D (Fig.~\ref{fig:generalization on adt3d}) further demonstrate the strong generalization capability of our method.

\subsection{Ablation Studies}

\noindent\textbf{Ablation of inference mode}
We evaluate our method under different inference modes on PointOdyssey3D. The “one” mode tracks a single query point, while the “all” mode simultaneously tracks all query points. In addition, we investigate the influence of incorporating auxiliary points. Specifically, KNN is used to introduce local neighboring points, whereas FPS and random sampling are adopted to provide globally distributed auxiliary points. By default, we add 1,024 auxiliary points.
As shown in Tab.~\ref{tab:inference_mode}, incorporating auxiliary points greatly improves performance. In particular, in the "one" mode, combining local (KNN) and global (FPS) auxiliary points achieves the best results, reducing EPE$_{3D}$ by 86\% over the baseline without auxiliary points and by 21.7\% compared with the variant that combines local (KNN) and global (random) auxiliary points. Notably, FPS-based global sampling yields greater accuracy gains than random sampling.

\begin{table}[t]
\centering
\resizebox{1\columnwidth}{!}{
\begin{tabular}{p{0.5cm}ccccc}
\toprule
\text{Query} & \text{Support} & $\text{EPE}_{3D}\textcolor{red}{\downarrow}$ & $\delta_{3D}^{\text{avg}}\textcolor{green!60!gray}{\uparrow}$ & $\text{Survival}_{3D}^{0.50}\textcolor{green!60!gray}{\uparrow}$ & $\text{MAE}_{3D}\textcolor{red}{\downarrow}$ \\ 
\midrule
One & - & 0.852 & 47.49 & 57.31 & 0.834 \\ 

One & Loc.KNN & 0.148 & 83.00 & 94.35 & 0.146 \\ 
One     & Glob.FPS & 0.134 & 87.64 & 95.05 & 0.136 \\ 
 One    & Glob.Random & 0.164 & 85.64 & 93.40 & 0.164 \\ 
 One    & \makecell{Loc.KNN \\ + Glob.Random} & 0.152 & 86.42 & 95.16 & 0.151 \\ 
 One    & \makecell{Loc.KNN \\ + Glob.FPS} & \textbf{0.119} & \textbf{87.64} & \textbf{95.74} & \textbf{0.115} \\ 
\midrule
All & - & 0.133 & 86.37 & 93.65 & 0.129 \\ 
  All  & Glob.Random & \textbf{0.124} & \textbf{87.94} & 95.01 & \textbf{0.118} \\ 
  All  & Glob.FPS & 0.125 & 87.93 & \textbf{95.03} & 0.119 \\ 
\bottomrule
\end{tabular}
}
\caption{ \textbf{Ablation of inference mode.} We evaluate different inference modes on the PointOdyssey3D dataset to analyze how various auxiliary point selection strategies affect performance.
}
\label{tab:inference_mode}
\end{table}

\noindent\textbf{Effectiveness of geometry feature update}
We assess the effect of geometry feature updating by removing the feature update branch. As shown in Tab.~\ref{tab: ablation} (a), excluding this component leads to a clear performance degradation, with $\text{EPE}_{3D}$ increasing by $34.2\%$. This suggests that, unlike pairwise scene flow estimation, long-term estimation must account for feature variations induced by geometric changes over time. Explicitly modeling such temporal geometric dynamics significantly enhances matching stability

\noindent\textbf{Impact of sliding window size}
We evaluate different sliding-window sizes to examine how the temporal context length affects performance. As shown in Tab.~\ref{tab: ablation} (b), extending the temporal context consistently improves accuracy and yields more reliable trajectory predictions. Compared with the two-frame setting, our model achieves notably lower errors, with $\text{EPE}_{3D}$ reduced by 35.4\%.

\begin{table}[!t]
	\centering
	
	\resizebox{1.0\columnwidth}{!}{%
		\begin{tabular}{@{}p{0.5cm}ccccc@{}}
			\toprule
			 &Experiment & Variation & $\text{EPE}_{3D}\textcolor{red}{\downarrow}$ & $\delta_{3D}^{avg}\textcolor{green!60!gray}{\uparrow}$ & $\text{Survival}_{3D}^{0.50}\textcolor{green!60!gray}{\uparrow}$  \\ \midrule
			
			\multirow{2}{*}{\centering (a)}
            &\multirow{2}{*}{\begin{tabular}[c]{@{}c@{}}Geometry–Motion \\joint Optimization\end{tabular}}
            & w/o & 0.202 & 75.85 & 90.61 \\
			& & w/ & \textbf{0.133} & \textbf{86.37} & \textbf{93.65} \\ \midrule
            \multirow{3}{*}{\begin{tabular}[c]{@{}c@{}}(b)\end{tabular}}
			&\multirow{3}{*}{\begin{tabular}[c]{@{}c@{}}Window  Size\end{tabular}} 
            & 2 & 0.206 & 78.33 & 90.68 \\
			&& 8 & 0.166 & 83.06 & 92.67 \\
			&& 16 &\textbf{0.133} & \textbf{86.37} & \textbf{93.65} \\ \midrule
            \multirow{3}{*}{\begin{tabular}[c]{@{}c@{}}(c)\end{tabular}}
			&\multirow{4}{*}{\begin{tabular}[c]{@{}c@{}}Spatial-Temporal \\Transformer\\Block's Number\end{tabular}} & 1 $\times$ 2 & 0.185 & 82.93 & 92.92 \\
            
            && 3 $\times$ 2 &\textbf{0.133} & 86.37 & \textbf{93.65} \\
            && 6 $\times$ 1 &{0.202} & 75.84 & {90.62} \\
			&& 6 $\times$ 2 & 0.140 & \textbf{86.50} & 93.29 \\ \bottomrule
		\end{tabular}%
	}
    \caption{\textbf{Results of ablation studies on the PointOdyssey3D.} }
	\label{tab: ablation}
\end{table}

\noindent\textbf{Impact of Transformer blocks' number}
We vary the number of Transformer blocks in the refinement stage to examine how layer depth affects accuracy. Excluding spatial dependencies (×1, temporal-only) leads to a significant degradation, whereas adopting $M=3$ spatial–temporal blocks (×2) yields the best $\text{EPE}_{3D}$ performance, which we use as the default setting in our model.

\begin{table}[!t]
\footnotesize
	\begin{center}
\resizebox{1.0\columnwidth}{!}
{  
\begin{tabular}{llcccc}  
\toprule 
\multirow{2}{*}{Dataset}       & \multirow{2}{*}{Method} & \multicolumn{4}{c}{Number of Frames}              \\ \cline{3-6}   
                               &                         & 2      & 8      & 24      & 40      \\ \hline  
\multirow{2}{*}{POD3D} & SF-baseline      & 0.0320 & 0.165 & 0.385 & 0.543 \\  
                        & Ours     & \bf0.0237 & \bf0.0719 & \bf0.157 & \bf0.205 \\ \hline  
\multirow{2}{*}{ADT3D} & SF-baseline  & 0.0363 & 0.1746 & 0.3999 & 0.5906 \\  
                        & Ours & \bf0.0199 & \bf0.0605 & \bf0.1261 & \bf0.1854 \\ \bottomrule 
\end{tabular}  
    }  

	\caption{\textbf{Comparison temporal drift across different frames.} POD3D denotes the PointOdyssey3D dataset.}	
	\label{table:frames}
	\end{center}
\end{table}

\noindent\textbf{Analysis of temporal drift}
We compare trajectory errors over time to analyze the accuracy evolution trend. As shown in Tab.~\ref{table:frames}, PCSTracker consistently outperforms SF-baseline at all timestamps and exhibits a slower error increase, indicating effective suppression of accumulation and drift.

\begin{table}[!t]
	\begin{center}
		\resizebox{0.8\columnwidth}{!}
		{
	   \begin{tabular}{lccc}
        \toprule
            Method & Parameters & Runtime &FPS \\
            \midrule
            SpatialTracker & 34.0M & 4.73s &8.46 \\
            SceneTracker & 24.2M & 1.38s &29.0\\
            Ours & 3.48M & 1.23s &32.5\\
            \bottomrule
        \end{tabular}
		}	
	
	\caption{\textbf{Parameters and runtime comparison.} Runtime comparison under the 40 frames, 1024 query points setting. }
	
	\label{table:time}
	\end{center}
\end{table}

\noindent\textbf{Parameters and runtime}
We compare SpatialTracker, SceneTracker, and our PCSTracker in terms of parameter count and runtime to show the efficiency of our method. All experiments run on a single RTX 3090 GPU. As shown in Tab.~\ref{table:time}, our model achieves both a smaller parameter size and faster runtime.

\subsection{Limitation}

As the model directly takes raw 3D point coordinates as input, it is sensitive to geometric scale and scene-dependent distance variations. This sensitivity is more pronounced when transferring from synthetic data (e.g., PointOdyssey) to real-world domains with distinct spatial distributions, such as autonomous driving. Future work will focus on introducing scene-specific data or adaptive training strategies to mitigate distribution shifts and enhance robustness and generalization in diverse real-world environments.
\section{Conclusion}
\label{sec:conclusion}

We propose PCSTracker, the first framework for long-term scene flow estimation in point clouds. By modeling geometric evolution, leveraging temporal context, and employing an overlapping sliding-window inference strategy, PCSTracker achieves consistent and fine-grained 3D motion estimation over point cloud sequences. We construct two datasets, PointOdyssey3D and ADT3D, for training and evaluation to further advance research on long-term scene flow in point clouds. Extensive experiments demonstrate that PCSTracker delivers superior temporal consistency, robustness, and 3D motion reasoning in dynamic scenes.

\textbf{Acknowledgement.} This research is supported by the National Key Research and Development Program of China (2024YFE0217700), the National Natural Science Foundation of China (62472184, 62122029), and the Fundamental Research Funds for the Central Universities.

{
    \small
    \bibliographystyle{ieeenat_fullname}
    \bibliography{main}
}

\clearpage
\clearpage
\setcounter{page}{1}
\maketitlesupplementary

\section{Number of Iteration}
\label{sec:iteration}

We vary the number of iterative refinement steps to evaluate their impact on overall performance.
As shown in Tab.~\ref{table:number_of_iter}, the model achieves its lowest $\text{EPE}_{3D}$ on the PointOdyssey3D dataset with K=6 iterations.
In practice, we use 
K=4 as the default setting, which offers a favorable trade-off between accuracy and computational efficiency.

\setlength{\tabcolsep}{1mm}
\begin{table}[h]
	\begin{center}
					
		\resizebox{1.00\columnwidth}{!}
            {
		\begin{tabular}{ccccccc}
		    \hline
			\toprule
			
		Dataset& Num. Iteration   & $\text{EPE}_{3D}\textcolor{red}{\downarrow}$ & $\delta_{3D}^{\text{avg}}\textcolor{green!60!gray}{\uparrow}$& $\text{Survival}_{3D}^{0.50}\textcolor{green!60!gray}{\uparrow}$& $\text{MAE}_{3D}\textcolor{red}{\downarrow}$\\ \midrule

			& 1    & 0.171 & 81.16 & 91.75 & 0.167\\			
			
			&  2       & 0.136  &  86.09 & 94.41  & 0.130 \\			
					
			&  4      & 0.133  &  86.37 & 93.65  & 0.129 \\

		    \multirow{-1}{*}{\begin{tabular}[c]{@{}c@{}}POD3D \end{tabular}}
			
			& 6    & \bf{0.124} & \bf{87.70} & \bf{95.05} & \bf{0.118}\\			
			
			&  8       & 0.126  &  87.52 & 95.02  & 0.120 \\			
					
			&  12      & 0.132  &  86.64 & 94.93  & 0.125 \\

			\midrule
			
			& 1    & 0.482 & 65.07 & 79.87 & 0.315\\			
			
			&  2       & 0.402  &  71.28 & 84.66  & 0.255 \\			
					
			&  4      & 0.372  &  74.44 & 87.74  & 0.226 \\
            
            \multirow{-3}{*}{\begin{tabular}[c]{@{}c@{}}ADT3D\end{tabular}}
            
            & 6    & 0.370 & \bf{74.72} & 88.42 & \bf{0.220}\\			
			
			&  8       & \bf{0.369}  &  74.16 & \bf{88.89}  &0.221 \\			
					
			&  12      & 0.381  &  71.08 & 88.15  & 0.232 \\
			\bottomrule
			\hline
		\end{tabular}
		}
	\vspace{-8pt}
	\caption{\textbf{Comparison results on PointOdyssey3D (POD3D) and ADT3D datasets with different iteration numbers.}}
	\label{table:number_of_iter}
	\end{center}
	\vspace{-6mm}
\end{table}

\section{Impact of Point-Voxel Correlation}
\label{sec:voxel scale}

We evaluate the effectiveness of the point–voxel dual-branch correlation pyramid by individually removing each branch, as shown in Tab.~\ref{table:correlation_compare}.
Removing the voxel correlation branch increases $\text{EPE}_{3D}$ by 73.5\%, whereas removing the point correlation branch results in an 11.3\% increase.

\section{Impact of the Number of Query Points}
\label{sec:Tracking Number}

Tab.~\ref{table:number_of_query_points} summarizes the effect of varying the number of query points on the final performance.
Experiments on both PointOdyssey3D and ADT3D show that accuracy is lowest when tracking a single query point. Increasing the number of query points from 1 to 32 results in  a substantial improvement, reducing $\text{EPE}_{3D}$ by 76.9\% and 53.9\%, respectively.
Further increasing the number of query points continues to provide stable performance improvements.
These results suggest that jointly estimating denser point sets leads to more stable and accurate motion predictions.

\setlength{\tabcolsep}{1.2mm}
\begin{table}[t]
	\begin{center}
					
		\resizebox{1.00\columnwidth}{!}
            {
		\begin{tabular}{ccccccc}
		    \hline
			\toprule
			
		Dataset&  Num. Query   & $\text{EPE}_{3D}\textcolor{red}{\downarrow}$ & $\delta_{3D}^{\text{avg}}\textcolor{green!60!gray}{\uparrow}$& $\text{Survival}_{3D}^{0.50}\textcolor{green!60!gray}{\uparrow}$& $\text{MAE}_{3D}\textcolor{red}{\downarrow}$\\ \midrule

			& 1    & 0.668 & 56.04 & 67.66 & 0.619\\			
			
			&  32       & 0.154  &  83.13 & 93.72  & 0.148 \\			
					
			&  64      & 0.143  &  85.24 & 93.99  & 0.137 \\

		    \multirow{-1}{*}{\begin{tabular}[c]{@{}c@{}}POD3D \end{tabular}}
			
			& 256    & 0.142 & 84.84 & 94.01 & 0.138\\			
			
			&  512       & 0.139  &  85.16 & \bf{94.34}  & 0.135 \\			
					
			&  1024      & \bf{0.133}  &  \bf{86.37} & 93.65  & \bf{0.129} \\

			\midrule
			
			& 1    & 1.279 & 37.86 & 53.02 & 0.969\\			
			
			&  32       & 0.589  &  59.94 & 75.08  & 0.394 \\			
					
			&  64      & 0.545  &  63.20 & 78.11  & 0.363 \\

            \multirow{-3}{*}{\begin{tabular}[c]{@{}c@{}}ADT3D\end{tabular}}
            
            & 256    & 0.501 & 66.25 & 80.81 & 0.327\\			
			
			&  512       & 0.489  &  66.43 & 80.96  & 0.318 \\			
					
			&  All      & \bf{0.372}  &  \bf{74.44} & \bf{87.74}  & \bf{0.226} \\
			\bottomrule
			\hline
		\end{tabular}
		}
	\caption{\textbf{Comparison results on PointOdyssey3D (POD3D) and ADT3D datasets with different numbers of query points.}}
	\label{table:number_of_query_points}
	\end{center}
	\vspace{-6mm}
\end{table}

\begin{table}[!t]
	\begin{center}
		\resizebox{1.0\columnwidth}{!}
		{
	   \begin{tabular}{ccccc}
        \toprule
            Correlation Volume & $\text{EPE}_{3D}\textcolor{red}{\downarrow}$ & $\delta_{3D}^{\text{avg}}\textcolor{green!60!gray}{\uparrow}$ &$\text{Survival}_{3D}^{0.50}\textcolor{green!60!gray}{\uparrow}$
            &$\text{MAE}_{3D}\textcolor{red}{\downarrow}$\\
            \midrule
            Point-only & 0.501 & 51.07 &65.16 &0.510 \\
            Voxel-only & 0.150 & 84.95 &93.22 &0.145\\
            Point-Voxel & \bf{0.133} & \bf{86.37} &\bf{93.65} &\bf{0.129}\\
            \bottomrule
        \end{tabular}
		}	
	
	\caption{\textbf{Impact of point-voxel correlation}  }

	\label{table:correlation_compare}
	\end{center}
\end{table}

\section{Robust for Occlusion}
\label{sec:occlision}

We compare the performance of PCSTracker with the SF-baseline under varying occlusion levels to evaluate their robustness to occlusions. As shown in Tab.~\ref{table:occ_robust}, our method consistently achieves substantially lower $\text{EPE}_{3D}$ on both POD3D and ADT3D across all occlusion levels, indicating significantly improved robustness. In addition, leveraging a broad temporal context not only benefits occluded points but also improves the accuracy of non-occluded points.

\begin{table}[h]
\footnotesize
	\begin{center}
\resizebox{1.0\columnwidth}{!}
{  
\begin{tabular}{llcccc}  
\toprule 
\multirow{2}{*}{Dataset}       & \multirow{2}{*}{Method} & \multicolumn{4}{c}{Number of Occlusion Frames}              \\ \cline{3-6}   
                               &                         & 0      & 2      & 4      & $>$ 4      \\ \hline  
\multirow{2}{*}{POD3D} & SF-baseline      & 0.1950 & 0.2421 & 0.2714 & 0.3511 \\  
                        & Ours     & \bf0.0754 & \bf0.1034 & \bf0.1200 & \bf0.1736 \\ \hline  
\multirow{2}{*}{ADT3D} & SF-baseline  & 0.3404 & 0.4188 & 0.4690 & 0.5233 \\  
                        & Ours & \bf0.0961 & \bf0.1228 & \bf0.1303 & \bf0.1464 \\ \bottomrule 
\end{tabular}  
    }  

	\caption{\textbf{Comparison performance under different occlusion levels.} POD3D denotes the PointOdyssey3D dataset.}	
	\label{table:occ_robust}
	\end{center}
\end{table}

\begin{figure*}[t]
    \centering
    \includegraphics[width=0.95\linewidth]{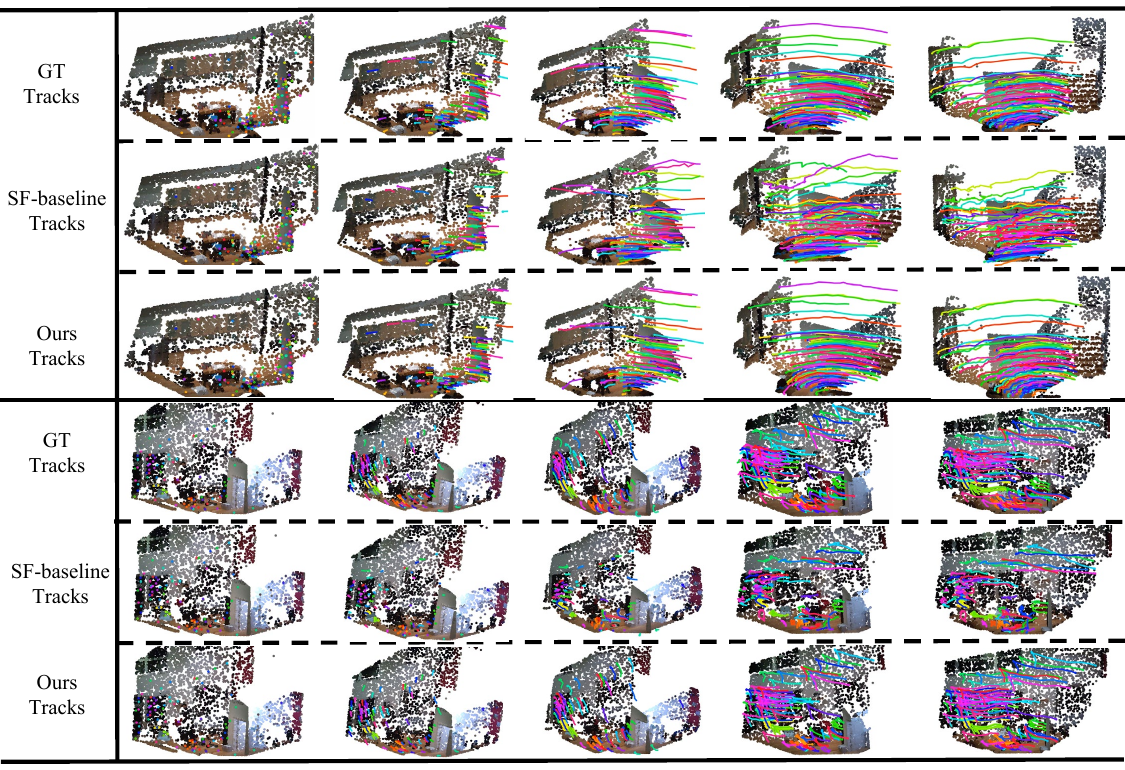}
    \caption{ \textbf{Appendix Results on ADT3D.} Each row shows the point clouds and the motion trajectories of the query points at different timestamps within a given scene (each point cloud is colored with its RGB information for better visualization).
    }
    \label{fig: appendix of adt3d1}
    \vspace{-10pt}
\end{figure*}

\section{Discussion of Limitation}
\label{sec:Limitation}

Although our method effectively estimates long-term scene flow directly from raw point-cloud sequences, it still exhibits an important limitation. The core challenge lies in its sensitivity to the distribution of input coordinates—particularly geometric scale and scene-dependent distance variations. When the distribution of test data deviates substantially from that of the training set, the model’s performance deteriorates sharply.

\begin{figure}[t]
    \centering
    \includegraphics[width=1.0\linewidth]{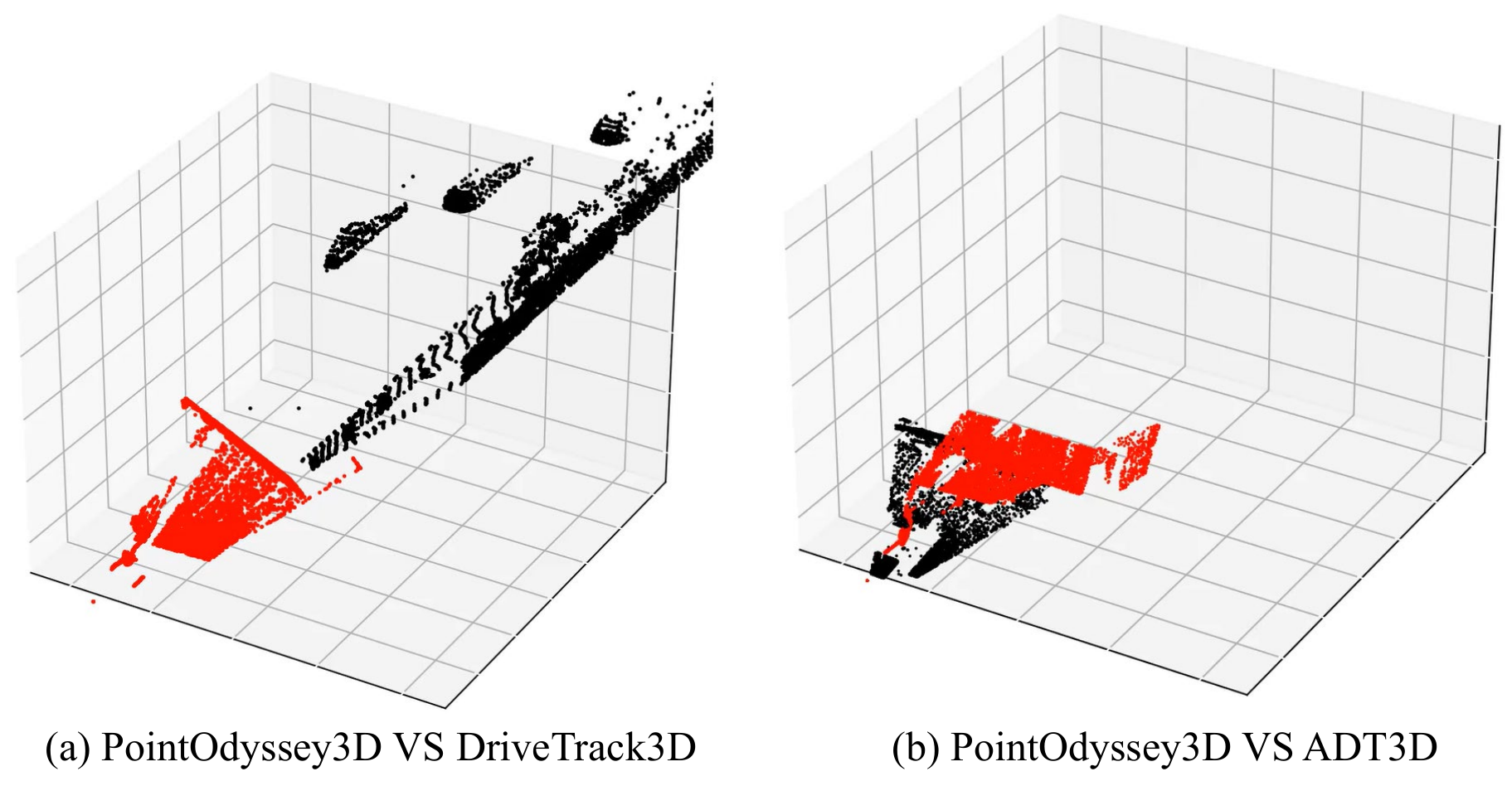}
    \caption{ \textbf{Spatial distribution gap across datasets.} \textcolor{red}{red} points denote the training dataset (POD3D), while black points represent the datasets used for generalization evaluation.  
    }
    \label{fig:dataset_gap}
    
\end{figure}

To illustrate this issue, we construct DriveTrack3D, an autonomous-driving dataset for long-term point cloud scene flow estimation, built from LiDAR sequences in the Waymo Open Dataset and trajectory annotations from TAPVid-3D. As shown in Tab. \ref{table: gene_drive}, our model fails entirely on DriveTrack3D, despite performing well on ADT3D datasets. Fig. \ref{fig:dataset_gap} further visualizes the spatial distribution gap: the training data and ADT3D share broadly similar geometric scales, whereas DriveTrack3D exhibits significantly larger depth ranges and outdoor spatial layouts.

This phenomenon raises an important research question for the community: how to achieve robust and generalizable long-term point cloud scene flow estimation across real-world environments with diverse spatial distributions and significant domain variations.

\begin{table}[h]
	\begin{center}
		\resizebox{0.6\columnwidth}{!}
		{
	   \begin{tabular}{lcc}
        \toprule
            Metric & DriveTrack3D & ADT3D  \\
            \midrule
            $\text{EPE}_{3D}(m)$ & 9.5 & 0.372  \\
            \bottomrule
        \end{tabular}
		}	
	
	\caption{\textbf{Generalization Results on Different Datasets.}}

	\label{table: gene_drive}
    \vspace{-10pt}
	\end{center}
\end{table}

\begin{figure*}[t]
    \centering
    \includegraphics[width=1.0\linewidth]{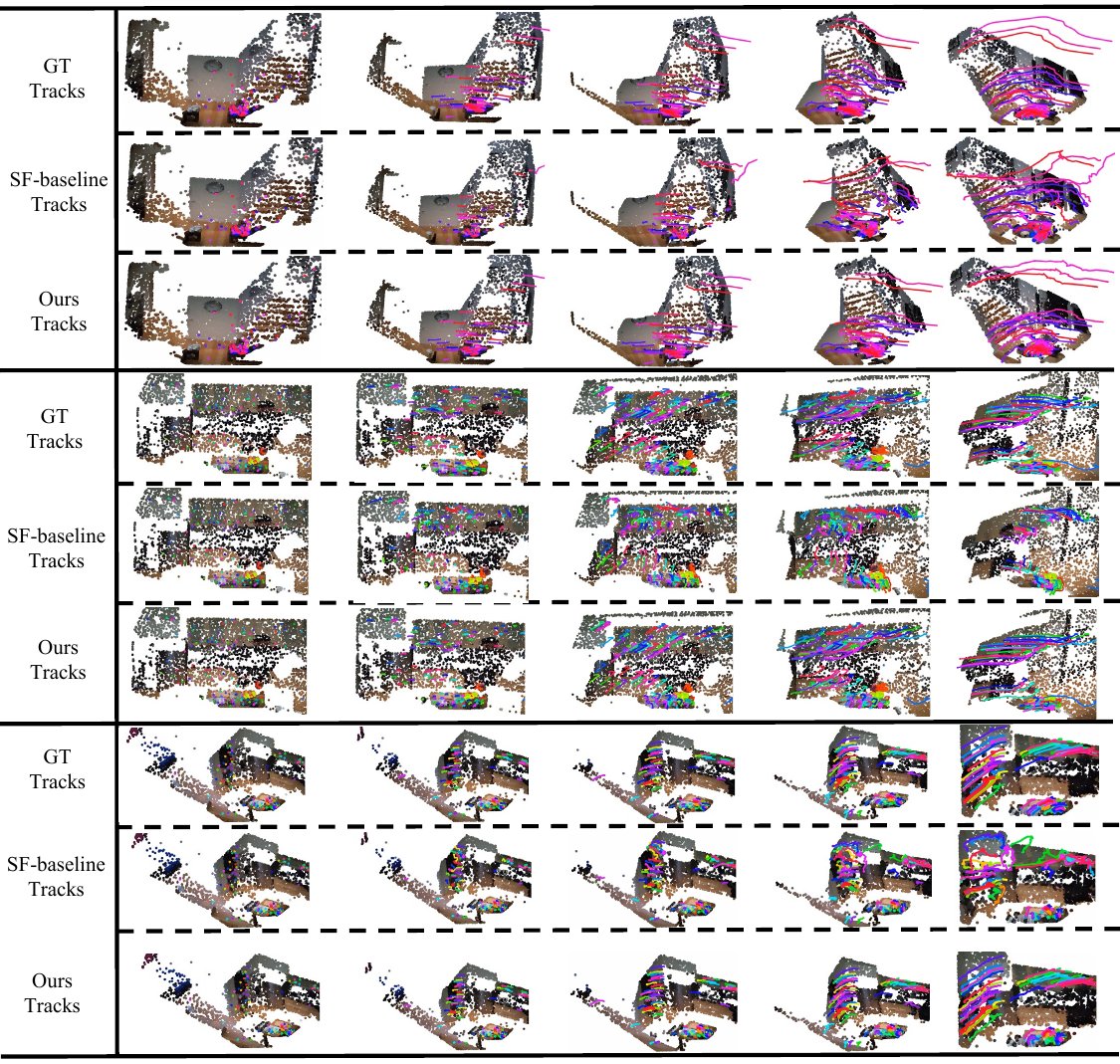}
    \caption{ \textbf{Appendix Results on ADT3D.} Each row shows the point clouds and the motion trajectories of the query points at different timestamps within a given scene (each point cloud is colored with its RGB information for better visualization).
    }
    \label{fig: appendix of adt3d2}
    \vspace{-10pt}
\end{figure*}

\begin{figure*}[t]
    \centering
    \includegraphics[width=1.0\linewidth]{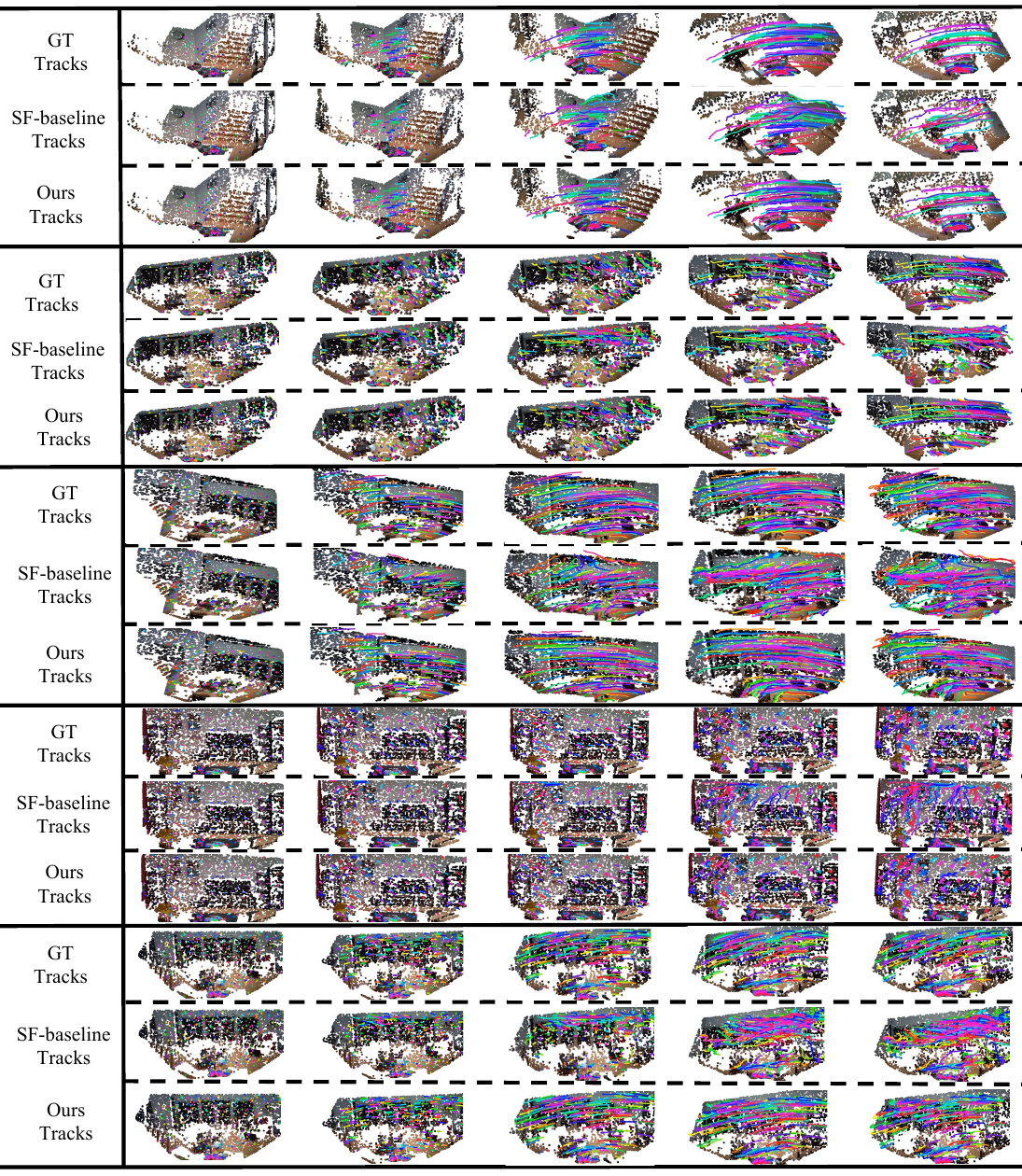}
    \caption{ \textbf{Appendix Results on ADT3D.} Each row shows the point clouds and the motion trajectories of the query points at different timestamps within a given scene (each point cloud is colored with its RGB information for better visualization). 
    }
    \label{fig: appendix of adt3d3}
    \vspace{-10pt}
\end{figure*}

\begin{figure*}[t]
    \centering
    \includegraphics[width=1.0\linewidth]{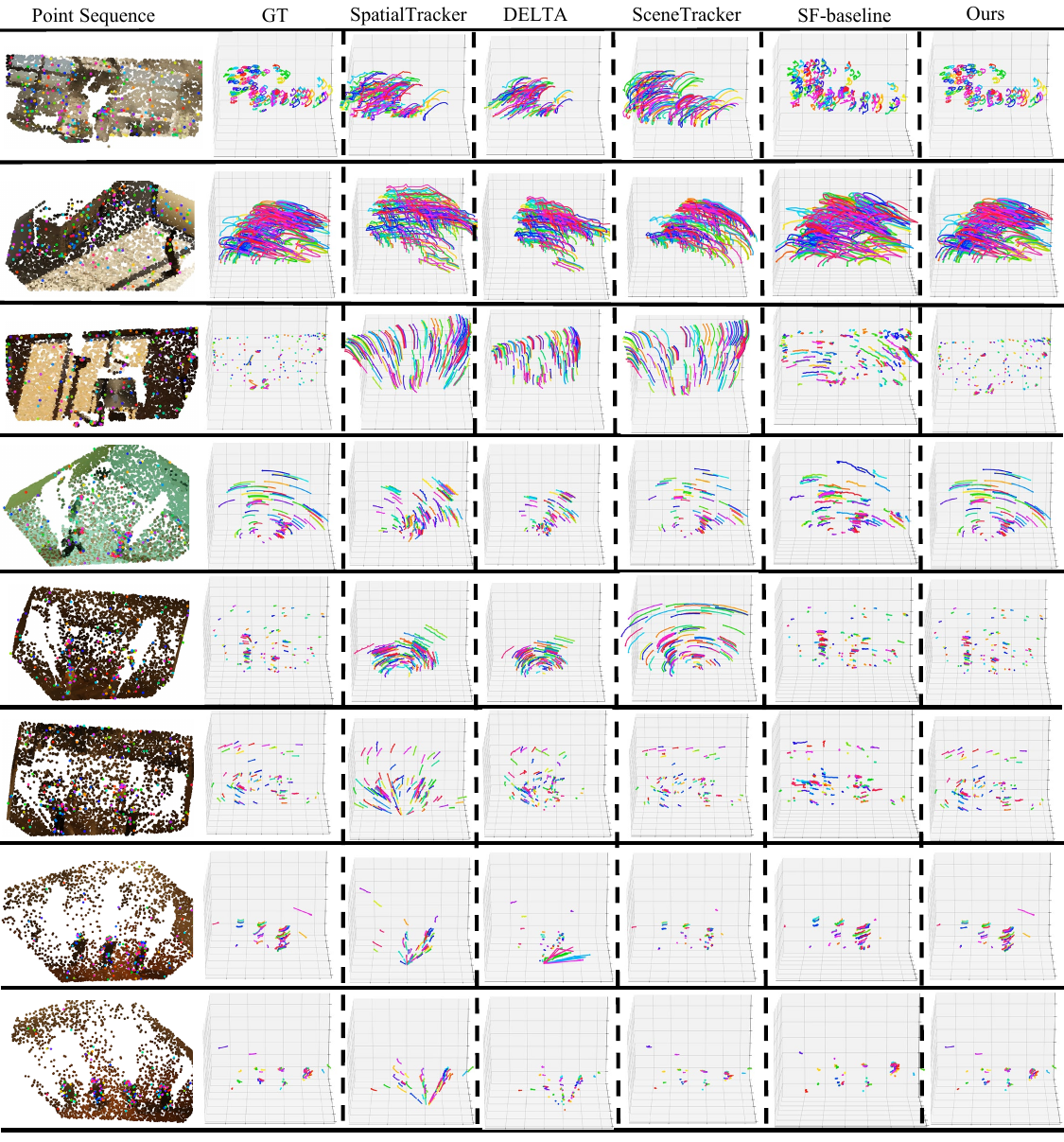}
    \caption{ \textbf{Appendix Results on PointOdyssey3D.} The first column shows the input point cloud sequence (colored by RGB for visualization) and the corresponding query points. Columns 2–7 compare ground-truth trajectories with predictions from different methods. 
    }
    \label{fig: appendix of pod3d1}
    \vspace{-10pt}
\end{figure*}

\section{Visualization}
\label{sec:visualization}

In this section, we present additional visualization results to further demonstrate the effectiveness of our method and its accurate understanding of 3D motion. Fig. \ref{fig: appendix of adt3d1}, \ref{fig: appendix of adt3d2}, and \ref{fig: appendix of adt3d3}  show more qualitative results on the ADT3D dataset, highlighting the strong generalization ability of our approach in real-world scenarios. Fig. \ref{fig: appendix of pod3d1} further compares trajectories predicted by different methods, showing that our model produces more accurate and geometrically consistent point-level 3D motion, thereby validating its effectiveness and reliability in 3D motion understanding.

\end{document}